%% file: 00_main.tex
\documentclass[letterpaper,10pt,conference]{ieeeconf}

\IEEEoverridecommandlockouts
\usepackage{cite}
\usepackage{amsmath,amssymb}
\usepackage{graphicx}
\usepackage{array}
\usepackage{xcolor}
\makeatletter
\let\NAT@parse\undefined
\makeatother
\usepackage{hyperref}
\usepackage{titlesec}

\makeatletter
\def\@IEEEauthorblockconfadjspace{-2em}
\makeatother

\begin{document}

\setlength{\abovecaptionskip}{2pt}
\setlength{\belowcaptionskip}{0pt}
\titlespacing*{\section}{0pt}{3pt}{1pt}
\titlespacing*{\subsection}{0pt}{2pt}{0.5pt}
\titlespacing*{\subsubsection}{0pt}{1pt}{0.5pt}

\setlength{\floatsep}{3pt plus 1pt minus 1pt}
\setlength{\textfloatsep}{3pt plus 1pt minus 1pt}
\setlength{\intextsep}{3pt plus 1pt minus 1pt}

% Applies the IEEEtran.bst control settings from `publications.bib`.
\bstctlcite{BSTcontrol}

% \title{Enhancing the Robustness of PINN-Based Power System State Estimation Under False Data Injection Attacks}
\title{\LARGE\bf
Learning Without Adversarial Training: A Physics-Informed Neural Network for Secure Power System State Estimation under False Data Injection Attacks
}
% \author{Solon Falas,~\IEEEmembership{Member,~IEEE}, Markos Asprou,~\IEEEmembership{Member,~IEEE}, Charalambos Konstantinou,~\IEEEmembership{Senior Member,~IEEE}, Maria K. Michael,~\IEEEmembership{Member,~IEEE}}
\author{Solon Falas, Markos Asprou, Charalambos Konstantinou, and Maria K. Michael%
\thanks{S. Falas and M. Asprou are with the KIOS Center of Excellence, University of Cyprus. C. Konstantinou is with CEMSE, King Abdullah University of Science and Technology (KAUST). M. K. Michael is with the KIOS Center of Excellence and the Department of Electrical and Computer Engineering, University of Cyprus.}%
}

\IEEEaftertitletext{\vspace{-2\baselineskip}}

\maketitle
\thispagestyle{empty}
\pagestyle{empty}

\begin{abstract}
Power System State Estimation (PSSE) converts geographically distributed measurements into the voltage magnitudes and phase angles needed for grid monitoring and control. Learned estimators can perform this mapping rapidly, but model-aware False Data Injection Attacks (FDIAs) may corrupt their inputs while retaining AC plausibility and residual-based stealth. Physics-Informed Neural Networks (PINNs) limit candidate states through power-flow consistency; however, their robustness depends on balancing supervised and physics losses whose scales and gradient contributions evolve during training. This paper proposes a PINN that jointly learns homoscedastic uncertainty parameters and uses them to adapt the two objectives. The formulation assigns trainable log-uncertainties to the active-power, reactive-power, voltage, and angle losses while safeguarding against an underweighted physics objective. The estimator is trained only on clean steady-state data and is evaluated, without adversarial retraining, under baseline state-distortion and stricter residual-profile-matching regimes on the IEEE~118-bus system. Accuracy is measured against the uncompromised system state. Relative to a fixed-weight PINN, dynamic weighting reduces overall Mean Absolute Error (MAE) by $55\%$ while also improving voltage- and angle-estimation accuracy. The results show that learning the physics/data balance from clean data improves robustness to unseen FDIAs.
\end{abstract}

% \vspace{-1\baselineskip}
% Paper sections (split across files in this repo).
\input{01_introduction.tex}
\input{02_related_work.tex}
\input{03_dynamic_weights.tex}
\input{04_fdia_construction.tex}
\input{05_experimental_setup.tex}
\input{06_conclusion.tex}

\bibliographystyle{IEEEtran}
\bibliography{publications,sources}

\end{document}

%% file: 01_introduction.tex
\section{Introduction}
Power System State Estimation (PSSE) is a core energy-management function that uses Supervisory Control and Data Acquisition (SCADA) and Phasor Measurement Unit (PMU) data to recover the voltage magnitudes and phase angles required for real-time monitoring, control, and security assessment~\cite{ding2020secure}. The conventional approach, Alternating-Current (AC) Weighted Least Squares (WLS) with residual-based Bad-Data Detection (BDD), iteratively solves a nonlinear estimation problem for every measurement snapshot, typically through Newton-type updates. Its performance therefore depends on an accurate network model, measurement redundancy, reliable initialization, and trustworthy measurements.

The increasing availability of synchronized grid measurements has motivated data-driven PSSE, where a trained neural estimator approximates the nonlinear measurement-to-state mapping through fast, single-pass inference rather than resolving the estimation problem for each snapshot. This benefit comes with a limitation: the learned mapping is shaped by the operating conditions seen during training, and inputs outside that distribution may yield inaccurate or physically inconsistent states even when nominal validation performance is strong. Covering every security-relevant condition during training is infeasible, since genuine attacks are rare, unsafe to stage on operational infrastructure, and combinatorial in location, access, strength, and intent. A security-oriented learned estimator should therefore remain dependable when measurement corruptions differ from anything observed during training.

Coordinated False Data Injection Attacks (FDIAs) are especially challenging in this setting. A model-aware adversary can alter selected measurements while accounting for the network topology and parameters, allowing the falsified snapshot to preserve AC feasibility and remain within conventional BDD residual limits~\cite{iranpour2024fdia,ran2026correlation}. The resulting state bias can evade standard screening while distorting voltage, angle, or line-flow awareness. Such errors may then propagate to downstream monitoring and control decisions, making estimator robustness a cyber-physical security requirement rather than only a question of predictive accuracy.

Physics-Informed Neural Networks (PINNs) reduce dependence on training examples by embedding physical relationships into the learning objective~\cite{raissi2019physics}. For PSSE, a supervised term promotes recovery of voltage magnitudes and phase angles, while a physics term penalizes disagreement with the AC power-flow equations~\cite{falas2025tii}. Physics information, however, does not provide robustness automatically: the supervised and physics losses may have different magnitudes and gradient contributions that shift throughout optimization~\cite{wang2021understanding}, so fixed coefficients may overemphasize one objective, require costly weight searches, and transfer poorly as stealth constraints change.

This work addresses that weakness through homoscedastic uncertainty weighting~\cite{cipolla2018multi}. Four trainable log-uncertainty parameters are optimized jointly with the network and turned into positive inverse-variance weights for the active-power, reactive-power, voltage, and angle losses. A one-sided group-level safeguard prevents the aggregate physics contribution from becoming underweighted without forcing a fixed physics/data ratio. The resulting balance can evolve during training rather than being predetermined.

The PINN is trained on non-attacked steady-state measurements and then tested, without adversarial retraining, under two model-aware, stealth-constrained AC-FDIA regimes: baseline state distortion and state distortion with stricter residual-profile matching. In both regimes, estimation error is measured against the uncompromised system state, so these post-training stress tests evaluate whether the learned physics/data balance transfers to measurement-integrity conditions absent from training. The contributions are:
\begin{itemize}
    \item An attack-agnostic PINN that adapts the balance between state-reconstruction accuracy and AC consistency during clean-only training.
    \item A unified AC-feasibility and stealth framework for benchmarking state-distortion attacks with and without residual-profile matching.
\end{itemize}
On the IEEE~118-bus benchmark, dynamic weighting reduces Mean Absolute Error (MAE) by $55\%$ relative to the fixed-weight PINN and by $72\%$ relative to the frozen-weight ablation. The latter comparison shows that robustness depends on adapting the weights throughout optimization rather than merely reusing the final learned values.

The remainder of this paper is organized as follows. Section~\ref{s:related_work} reviews related work. Sections~\ref{s:dynamic_weights} and~\ref{ss:attack_construct} present the uncertainty-weighted estimator and attack framework. Section~\ref{s:exp_results} reports the IEEE~118-bus evaluation, and Section~\ref{s:conclusion} concludes.

%% file: 02_related_work.tex
\section{Related Work}\label{s:related_work}
Classical PSSE commonly uses an AC WLS formulation with residual-based bad-data processing. While these methods are well understood and widely deployed, performance can degrade with imperfect models, limited measurement redundancy, and structured bad data that simple bad data detection mechanisms do not capture well~\cite{ding2020secure,jin2019cdc}. Robust variants (e.g., alternative loss functions and screening heuristics) improve tolerance to random outliers but do not fully address coordinated adversarial manipulation.

FDIA research shows that attackers can construct stealthy attacks with respect to bad data detection mechanisms while still inducing targeted bias in the estimated state and derived quantities such as line flows~\cite{deng2016false}. This motivates estimator-side defenses to incorporate additional defence layers beyond the traditional bad data detection mechanisms that are based on residual tests, including physics consistency and checks that reduce the feasible space of stealthy corruptions.

Recent machine learning approaches for PSSE and related inference tasks span physics-guided objectives, physics-informed constraints, and topology-aware architectures. One line of work unrolls classical solvers into trainable networks, for example, by mapping Gauss-Newton iterations to layers and learning step sizes or priors for improved convergence and robustness~\cite{zhang2019real,yang2022data}.

A complementary strand adds physics to the learning objective. Under limited observability, physics-aware models embed AC power-flow relationships as residual penalties aiming toward physically consistent solutions~\cite{ngo2024physics}. Related formulations add Kirchhoff or power-flow constraints as soft regularizers and report improvements under noise and bad data~\cite{tran2021enhancement}. Physics-guided residual learning similarly augments data-driven estimators with physics-based correction terms~\cite{wang2020physicsguided}. Topology-aware variants incorporate network structure explicitly, including physics-informed graph neural networks aligned with bus-branch structure, as well as hybrid predictors coupled with physical simulators~\cite{ngo2024physics,tian2020hybrid}.

A recurring practical issue concerns balancing supervised data-fit terms and physics-residual terms when magnitudes vary across operating points and disturbance regimes. PINN-based PSSE is therefore a multiobjective training problem with heterogeneous but coupled terms. Although the model serves a single end task, training jointly optimizes state-recovery losses on $(\mathbf{V},\boldsymbol{\theta})$ and physics-consistency losses on quantities reconstructed from the same predictions. Accordingly, training can be cast as a multitask optimization problem and handled with uncertainty-based weighting~\cite{cipolla2018multi}. This provides a lightweight mechanism to learn the relative scaling of correlated objectives during training, rather than relying on manual weight sweeps.

Despite this, most PINN-based PSSE formulations still use fixed loss weights tuned offline~\cite{falas2025tii}, which becomes brittle under adversarial distribution shifts. The focus is not only robustness under these AC FDIAs, but whether adaptive weighting maintains an effective physics/data balance under stricter stealth. The evaluation uses uncertainty-based dynamic weighting and tests state-distortion attacks with and without residual-profile matching.

%% file: 03_dynamic_weights.tex
\section{Uncertainty-Weighted PINN}
\label{s:dynamic_weights}

% A dynamic loss-weighting scheme based on homoscedastic uncertainty in the PINN learns the physics vs. data tradeoff during training. The formulation couples per-metric uncertainty weighting with a one-sided ratio regularizer on aggregate physics-vs-data inverse-variance weights. The objective provides flexible and adaptive loss-parameter weighting while discouraging physics collapse during training.

A dynamic loss-weighting scheme based on homoscedastic uncertainty is proposed to learn the balance between measurement fitting and power-system physics during training. The formulation couples per-component uncertainty weighting with a one-sided ratio regularizer that prevents the aggregate physics contribution from becoming underweighted.

\begin{table}[h]
    \centering
    \textbf{Nomenclature}
    \vspace{0.5\baselineskip}

    \resizebox{\columnwidth}{!}{
    \begin{tabular}{ll}
        \hline
        Symbol & Description \\
        \hline
        $n$ & sample index in a minibatch, $n = 1,\dots,N_b$ \\
        $\mathbf{y}^{(n)}$ & input features for sample $n$ \\
        $(\mathbf{P}^{(n)},\mathbf{Q}^{(n)})$ & measured injection inputs for sample $n$ \\
        $(\mathbf{V}^{(n)},\boldsymbol{\theta}^{(n)})$ & supervised state targets for sample $n$ \\
        $(\hat{\mathbf{P}}^{(n)},\hat{\mathbf{Q}}^{(n)},\hat{\mathbf{V}}^{(n)},\hat{\boldsymbol{\theta}}^{(n)})$ & model outputs for sample $n$ \\
        $(\hat{\mathbf{P}}^{\mathrm{inj},(n)},\hat{\mathbf{Q}}^{\mathrm{inj},(n)})$ & injections reconstructed from predicted states \\
        $G_{ij},B_{ij}$ & network conductance/susceptance entries for buses $i,j$ \\
        $m \in \{p,q,v,\theta\}$ & loss-component index \\
        \hline
    \end{tabular}
    }
\end{table}

The Neural Network (NN) architecture, input-output parameterization, and physics residual definitions follow this PINN state-estimation setup: the model maps active/reactive power injections $\mathbf{y}=[\mathbf{P},\mathbf{Q}]$ to joint estimates $(\hat{\mathbf{P}},\hat{\mathbf{Q}},\hat{\mathbf{V}},\hat{\boldsymbol{\theta}})$ using a fully connected network of fixed depth and width, and the physics loss enforces AC net power injections via differentiable residuals.

The following equations define the dynamic weighting mechanism.
For each sample $n$, the model maps measured net powers to joint state and injection estimates:
\begin{equation}
\resizebox{0.9\columnwidth}{!}{$
\mathbf{y}^{(n)} = \big[\mathbf{P}^{(n)},\mathbf{Q}^{(n)}\big] \in \mathbb{R}^{2N_{\text{bus}}}, \quad
(\hat{\mathbf{P}}^{(n)},\hat{\mathbf{Q}}^{(n)},\hat{\mathbf{V}}^{(n)},\hat{\boldsymbol{\theta}}^{(n)}) = f_{\Theta}(\mathbf{y}^{(n)})
$}
\end{equation}
where $f_{\Theta}$ is the neural network with learnable parameters $\Theta$. This mirrors realistic SCADA-style availability where power injections are directly measured while states remain latent. The measured injections act as noisy inputs, while the supervised state targets remain $(\mathbf{V}^{(n)},\boldsymbol{\theta}^{(n)})$.
Predicting denoised $(\hat{\mathbf{P}},\hat{\mathbf{Q}})$ alongside $(\hat{\mathbf{V}},\hat{\boldsymbol{\theta}})$ means both supervised state and physics terms are evaluated on network outputs. This avoids matching reconstructed injections directly to noisy input measurements, which would create a circular output-to-input dependency. Instead, the model jointly learns state estimation and injection denoising; the physics loss compares predicted injections $(\hat{\mathbf{P}},\hat{\mathbf{Q}})$ with AC-consistent reconstructions from $(\hat{\mathbf{V}},\hat{\boldsymbol{\theta}})$.

The model computes physics residuals by reconstructing net power injections from predicted voltage magnitudes and angles using the AC power-injection equations. Given $\hat{\mathbf{V}}$ and $\hat{\boldsymbol{\theta}}$, AC power-injection reconstructions follow:
\begin{equation}
\resizebox{0.9\columnwidth}{!}{$
\hat{P}^{\mathrm{inj},(n)}_i = \sum_{j=1}^{N_{\text{bus}}} \hat{V}^{(n)}_i\hat{V}^{(n)}_j\left(G_{ij}\cos(\hat{\theta}^{(n)}_i - \hat{\theta}^{(n)}_j) + B_{ij}\sin(\hat{\theta}^{(n)}_i - \hat{\theta}^{(n)}_j)\right)
$}
\label{eq:ac_pinj}
\end{equation}
\begin{equation}
\resizebox{0.9\columnwidth}{!}{$
\hat{Q}^{\mathrm{inj},(n)}_i = \sum_{j=1}^{N_{\text{bus}}} \hat{V}^{(n)}_i\hat{V}^{(n)}_j\left(G_{ij}\sin(\hat{\theta}^{(n)}_i - \hat{\theta}^{(n)}_j) - B_{ij}\cos(\hat{\theta}^{(n)}_i - \hat{\theta}^{(n)}_j)\right)
$}
\label{eq:ac_qinj}
\end{equation}
The physics loss compares $(\hat{\mathbf{P}},\hat{\mathbf{Q}})$ with $(\hat{\mathbf{P}}^{\mathrm{inj}},\hat{\mathbf{Q}}^{\mathrm{inj}})$, preserving differentiability and AC consistency.

Each metric uses normalized loss for scale robustness:
\begin{equation}
\resizebox{0.9\columnwidth}{!}{$
    \mathcal{L}_{\mathrm{norm}}(a,b) = \frac{1}{d}\sum_{k=1}^{d}\left(\frac{a_k - \mu_a}{\sigma_a + \varepsilon} - \frac{b_k - \mu_a}{\sigma_a + \varepsilon}\right)^2 = \frac{1}{d}\sum_{k=1}^{d}\left(\frac{a_k - b_k}{\sigma_a + \varepsilon}\right)^2
$}
\end{equation}
where $(\mu_a,\sigma_a)$ are minibatch moments of the reference quantity $a$, $b$ denotes the comparison quantity, and $d$ is the vector dimension. Statistics from the reference quantity $a$ keep normalization anchored to the physical scale of each target and avoid arbitrary manual rescaling between heterogeneous variables. In practice, this normalization remained stable because training uses low-variance steady-state samples and adequately large minibatches.

Finally, the four loss components are:
\begin{equation}
\resizebox{0.9\columnwidth}{!}{$
\mathcal{L}_{p} = \frac{1}{N_b}\sum_{n=1}^{N_b}\mathcal{L}_{\mathrm{norm}}\!\left(\hat{\mathbf{P}}^{(n)},\hat{\mathbf{P}}^{\mathrm{inj},(n)}\right),\quad
\mathcal{L}_{q} = \frac{1}{N_b}\sum_{n=1}^{N_b}\mathcal{L}_{\mathrm{norm}}\!\left(\hat{\mathbf{Q}}^{(n)},\hat{\mathbf{Q}}^{\mathrm{inj},(n)}\right)
$}
\end{equation}
\begin{equation}
\resizebox{0.9\columnwidth}{!}{$
\mathcal{L}_{v} = \frac{1}{N_b}\sum_{n=1}^{N_b}\mathcal{L}_{\mathrm{norm}}\!\left(\mathbf{V}^{(n)},\hat{\mathbf{V}}^{(n)}\right),\quad
\mathcal{L}_{\theta} = \frac{1}{N_b}\sum_{n=1}^{N_b}\mathcal{L}_{\mathrm{norm}}\!\left(\boldsymbol{\theta}^{(n)},\hat{\boldsymbol{\theta}}^{(n)}\right)
$}
\end{equation}
Batch means preserve scale comparability across components. The loss has three steps:

\textbf{Step 1: Log-uncertainty parameterization.} Each component loss gets a trainable clipped log-uncertainty weight:
\begin{equation}
\resizebox{0.9\columnwidth}{!}{$
s^{\mathrm{clip}}_m = \mathrm{clip}(s_m,s_{\min},s_{\max}) \quad
\mathcal{J}_{\mathrm{dyn}} = \sum_{m}\left(\frac{1}{2}e^{-2s^{\mathrm{clip}}_m}\mathcal{L}_m + s^{\mathrm{clip}}_m\right)
$}
\end{equation}
Here, $s_m=\log \sigma_m$ is a trainable per-component log-uncertainty, $s_m^{\mathrm{clip}}$ is its bounded version, and $e^{-2s_m^{\mathrm{clip}}}$ is the corresponding inverse-variance weight. The fixed bounds $[s_{\min},s_{\max}]$ keep the learned scales well conditioned and prevent domination or collapse of any single loss term.

The inverse-variance factors $w_m = e^{-2s^{\mathrm{clip}}_m}$ set each component's relative emphasis. Role-wise aggregation gives:
\begin{equation}
W_{\mathrm{phys}} = \sum_{m\in\{p,q\}} w_m,\quad W_{\mathrm{data}} = \sum_{m\in\{v,\theta\}} w_m
\end{equation}
Role-wise aggregation regularizes at the group level (physics vs.\ data) without removing per-metric adaptivity.

\textbf{Step 2: Physics vs. data balance measurement.} The current balance is measured through:
\begin{equation}
r = \frac{W_{\mathrm{phys}}}{W_{\mathrm{data}} + \varepsilon}\quad
r_{\star} = \frac{n_{\mathrm{phys}}}{n_{\mathrm{data}}}
\end{equation}
where $n_{\mathrm{phys}}$ and $n_{\mathrm{data}}$ denote the number of physics and data components in the loss function, respectively. In this setup, $n_{\mathrm{phys}}=n_{\mathrm{data}}=2$, so $r_{\star}=1$. This neutral reference reflects equal component counts and does not require learned aggregate weights to converge to $1{:}1$.

\textbf{Step 3: One-sided log-ratio safeguard.} The penalty applies only when physics influence falls below target:
\begin{equation}
\resizebox{0.875\columnwidth}{!}{$
\Delta = \log(r_{\star} + \varepsilon) - \log(r + \varepsilon) \quad
\mathcal{P}_{\mathrm{ratio}} = \lambda_r\,[\max(0,\Delta)]^2
$}
\end{equation}
The hinge form keeps training unconstrained when physics influence is sufficient, and the log domain penalizes relative imbalance rather than absolute differences. The coefficient $\lambda_r$ sets how strongly low-physics regimes are discouraged, so the penalty acts as a soft safeguard against physics collapse rather than a hard equality constraint at convergence.

The finalized training objective becomes $\mathcal{L}_{\mathrm{final}} = \mathcal{J}_{\mathrm{dyn}} + \mathcal{P}_{\mathrm{ratio}}$, combining the uncertainty-weighted components with the ratio safeguard.

Overall, the design combines two complementary principles: (i) uncertainty-based local adaptation across individual losses and (ii) a minimal global safeguard on the physics-vs-data balance. This makes the objective interpretable (each term has a distinct role), nonarbitrary (a statistical or structural rationale supports each coefficient), and practically robust to changing operating conditions and attack strengths.

%% file: 04_fdia_construction.tex
\section{Stealth-Constrained AC-FDIA Design}
\label{ss:attack_construct}

This section introduces a systematic optimization pipeline for generating AC FDIA benchmarks to evaluate dynamic-PINN robustness. The threat model assumes a \emph{stealthy, model-aware} adversary who can manipulate a subset of measurements \emph{in transit} on the utility Wide Area Network (WAN). The attacker knows (or accurately approximates~\cite{deng2016false}) the estimator's network model (topology and parameters) and solves constrained attack problems to craft bounded perturbations that remain physically plausible under AC power-flow constraints while satisfying empirical residual-threshold stealth criteria motivated by residual-based BDD~\cite{gao2015cdc}.

Accordingly, the generated attacks are AC-consistent and residual-constrained, while preserving key external-grid interactions and remaining within specified residual thresholds. This systematic design yields controlled, repeatable benchmark scenarios for stress-testing estimator robustness, in contrast to unconstrained perturbation models (e.g., additive noise, naive scaling, random outliers) and availability attacks (e.g., Denial of Service (DoS)).

Within this scope, the evaluation uses two representative optimization-based FDIA regimes under a common attacker capability set: (i) baseline state distortion and (ii) state distortion with stricter residual-profile matching. Both regimes maximize the same state-deviation objective and use a unified constrained-optimization template with a common feasible set (AC consistency, per-bus residual thresholds, boundary-transfer preservation, regional power conservation, and operational limits). The second regime adds residual-profile-matching constraints, yielding controlled, repeatable attack samples under two levels of stealth restriction.

\begin{figure*}[t]
    \centering
    \includegraphics[width=\textwidth]{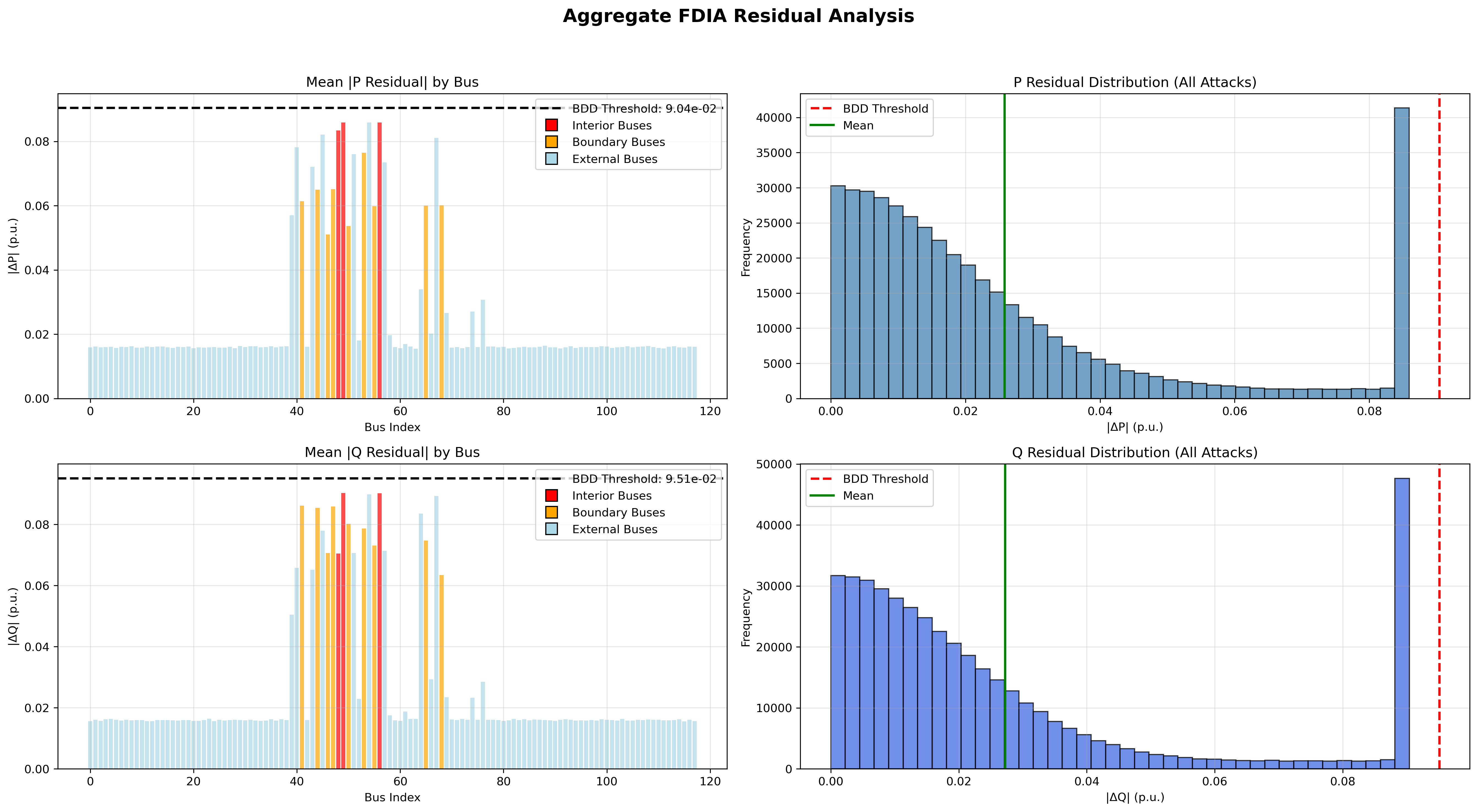}
    \vspace{-2\baselineskip}
    \caption[Simple FDIA Attack on IEEE~118-bus system]{Simple FDIA attack on the IEEE~118-bus system (Zone~2). Left: mean absolute active- and reactive-power residuals by bus. Right: distributions of residual magnitudes over all attacked snapshots.
    }
    \label{fig:simple_zone13}
    \vspace{-\baselineskip}
\end{figure*}

\subsection{Unified Stealth and Design Constraints}

The pipeline starts from uncorrupted steady-state snapshots and constructs attacked states in a connected topological zone $\mathcal{Z}$. The baseline and attacked tuples are $(\mathbf{P}^0,\mathbf{Q}^0,\mathbf{V}^0,\boldsymbol{\theta}^0)$ and $(\mathbf{P}^a,\mathbf{Q}^a,\mathbf{V}^a,\boldsymbol{\theta}^a)$, respectively.
Candidate zones are generated by breadth-first search on the grid graph, where $A_{ij}>0$ indicates adjacency between buses $i$ and $j$. Starting from a seed bus, the search uses hop limit $h_{\max}$ and size bounds $n_{\min},n_{\max}$ while optional radial expansion controls final attack-surface size.

For a selected zone $\mathcal{Z}$, buses are partitioned into interior and boundary sets, $\mathcal{B}_{\mathrm{int}}$ and $\mathcal{B}_{\mathrm{bnd}}$, with $\mathcal{Z}=\mathcal{B}_{\mathrm{int}}\cup\mathcal{B}_{\mathrm{bnd}}$. Buses outside $\mathcal{Z}$ are treated as exterior and fixed to baseline values.
The zone zero-injection set $\mathcal{B}_{\mathrm{zi}}$ follows from clean measurements via $|P_i^0|,|Q_i^0|<10^{-6}$. Interior buses are the main locations for injection and state changes, while boundary buses preserve aggregate external transfer.

AC-consistency constraints enforce nonlinear network relations at each bus $i$. The AC-reconstructed injections $P_i^{\mathrm{inj},a},Q_i^{\mathrm{inj},a}$ follow the power-injection equations~\eqref{eq:ac_pinj}--\eqref{eq:ac_qinj}, evaluated on the attacked state $(\mathbf{V}^a,\boldsymbol{\theta}^a)$ (superscript $a$ denotes attacked quantities).

The stealthiness is ensured through a residual-threshold:
\begin{equation}
\tau_P = 0.95\bar{\tau}_P,\quad \tau_Q = 0.95\bar{\tau}_Q
\end{equation}
where $\bar{\tau}_P,\bar{\tau}_Q$ denote residual scales computed as the maximum residual magnitudes over a steady-state dataset under normal operating conditions. The resulting constraints are:
\begin{equation}
|P_i^a-P_i^{\mathrm{inj},a}| \leq \tau_P,\quad |Q_i^a-Q_i^{\mathrm{inj},a}| \leq \tau_Q
\end{equation}

Boundary-transfer preservation applies per boundary bus on active flow to adjacent exterior buses. Let $F_{\mathrm{bnd},i}^{a}$ and $F_{\mathrm{bnd},i}^{0}$ denote attacked and baseline boundary active transfer at bus $i$. The constraint for each $i\in\mathcal{B}_{\mathrm{bnd}}$ is
\begin{equation}
\left|F_{\mathrm{bnd},i}^a-F_{\mathrm{bnd},i}^0\right|
\leq
\max\left(\varepsilon_{\mathrm{bnd,rel}}\left|F_{\mathrm{bnd},i}^0\right|,\varepsilon_{\mathrm{bnd,abs}}\right)
\label{eq:bnd_transfer_tolerance}
\end{equation}
Here, \textit{bnd} denotes boundary-interface quantities, $\varepsilon_{\mathrm{bnd,rel}}$ is the relative tolerance factor, and $\varepsilon_{\mathrm{bnd,abs}}$ is the absolute minimum tolerance. Active power captures net interchange and line-loading signatures with the external grid.

Regional power balance keeps net active and reactive injections close to the initial zonal totals. The implementation uses the shared conservation tolerance:
\begin{equation}
\resizebox{0.875\columnwidth}{!}{$
\begin{gathered}
\left|\sum_{i\in\mathcal{Z}}(P_i^a-P_i^0)\right| \leq \varepsilon_{\mathrm{cons}}, \quad
\left|\sum_{i\in\mathcal{Z}}(Q_i^a-Q_i^0)\right| \leq \varepsilon_{\mathrm{cons}} \\
\varepsilon_{\mathrm{cons}} = \max\left(10^{-3}\left|\sum_{i\in\mathcal{Z}}P_i^0\right|,10^{-3}\right)
\end{gathered}
$}
\end{equation}

Operational limits constrain $V_i^a$ to $[V_{\min},V_{\max}]$ and $\theta_i^a$ to $[\theta_{\min},\theta_{\max}]$. For each $i\in\mathcal{B}_{\mathrm{zi}}$, the zero-injection constraints enforce $P_i^a = 0,\quad Q_i^a = 0$.

For each sample, the attacker computes attacked variables by solving an attacker-side worst-case optimization problem that maximizes a reference-relative attack impact score:
\begin{equation}
\scalebox{0.95}{$
\begin{gathered}
\max \quad f_{\mathrm{attack}}(\mathbf{P}^a,\mathbf{Q}^a,\mathbf{V}^a,\boldsymbol{\theta}^a) \\
\text{over } \Delta \mathbf{P}^{|0,a|},\Delta \mathbf{Q}^{|0,a|},\Delta \mathbf{V}^{|0,a|},\Delta \boldsymbol{\theta}^{|0,a|} \\
\text{s.t.} \quad (\mathbf{P}^a,\mathbf{Q}^a,\mathbf{V}^a,\boldsymbol{\theta}^a)\in\mathcal{F}_{\mathrm{shared}}
\end{gathered}
$}
\end{equation}
where $\mathcal{F}_{\mathrm{shared}}$ denotes the shared feasible set. The maximization operator selects the feasible attacked state that produces the largest deviation. The two regimes keep the attacker capability assumptions and state-distortion objective fixed, while the residual-profile-matching regime adds the constraints defined below.

\subsection{Simple FDIA (Baseline State Distortion)}

The baseline regime maximizes voltage-magnitude and phase-angle deviations within attack zone $\mathcal{Z}$. The active- and reactive-injection changes at buses $i\in\mathcal{B}_{\mathrm{int}}\cup\mathcal{B}_{\mathrm{bnd}}$ are bounded through the shared constraints:
\begin{equation}
\resizebox{0.875\columnwidth}{!}{$
\max\; f_{\mathrm{attack}}(\mathbf{P}^a,\mathbf{Q}^a,\mathbf{V}^a,\boldsymbol{\theta}^a)
=
\sum_{i\in\mathcal{Z}} (V_i^a-V_i^0)^2
+
\sum_{i\in\mathcal{Z}} (\theta_i^a-\theta_i^0)^2
$}
\label{eq:fattack_obj}
\end{equation}
For each bus $i\in\mathcal{Z}$, injection bounds are written in the same absolute-deviation form used in Eq.~\eqref{eq:bnd_transfer_tolerance}:
\begin{equation}
\resizebox{0.875\columnwidth}{!}{$
\left|P_i^a-P_i^0\right| \leq \kappa_P\max(|P_i^0|,\delta_P),\quad
\left|Q_i^a-Q_i^0\right| \leq \kappa_Q\max(|Q_i^0|,\delta_Q)
$}
\end{equation}
where $\kappa_P,\kappa_Q<1$ scale the envelope widths and $\delta_P,\delta_Q$ set minimum thresholds. Exterior and zero-injection buses are fixed by the shared feasible set.
The attack, therefore, produces a false but operationally plausible state inside $\mathcal{Z}$ while keeping the shared stealth, balance, and operational constraints active, as seen in Fig.~\ref{fig:simple_zone13}. This scenario tests whether the estimator remains close to the uncompromised state when its inputs are consistent with an alternative state.
In the benchmark suite, this scenario serves as the baseline state-distortion regime and provides a reference for comparison with stricter residual-profile matching.

\subsection{State Estimation Corruption Attack (Residual-Matched)}

This regime maximizes the same state deviation while matching residual patterns to baseline values. Let the clean residuals be defined from the baseline snapshot as:
\begin{equation}
r_{P,i}^0 = P_i^0 - P_i^{\mathrm{inj},0},\quad r_{Q,i}^0 = Q_i^0 - Q_i^{\mathrm{inj},0}
\end{equation}
where $P_i^{\mathrm{inj},0}$ and $Q_i^{\mathrm{inj},0}$ are the AC reconstructions obtained from $(\mathbf{V}^0,\boldsymbol{\theta}^0)$. For each bus $i$, the constraints enforce:
\begin{equation}
\resizebox{0.85\columnwidth}{!}{$
\left| (P_i^a-P_i^{\mathrm{inj},a})-r_{P,i}^0 \right| \leq \delta_{P,i},\quad \delta_{P,i}=\max\left(\beta\left|r_{P,i}^0\right|,\varepsilon_r\right)
$}
\end{equation}
\begin{equation}
\resizebox{0.85\columnwidth}{!}{$
\left| (Q_i^a-Q_i^{\mathrm{inj},a})-r_{Q,i}^0 \right| \leq \delta_{Q,i},\quad \delta_{Q,i}=\max\left(\beta\left|r_{Q,i}^0\right|,\varepsilon_r\right)
$}
\end{equation}
The attack objective remains the zone-level state-distortion of Eq.~\eqref{eq:fattack_obj}, where $\beta$ is a relative residual-matching tolerance and $\varepsilon_r$ provides a nonzero tolerance floor when baseline residuals are small. This yields a stringent residual-profile-matching benchmark under the same AC and zonal constraints.
This setting captures the stricter stealth condition, preserving residual signatures while still trying to maximize state-estimation error.

%% file: 05_experimental_setup.tex
\begin{figure*}[ht]
    \vspace{-0.5\baselineskip}
    \centering
    \includegraphics[width=0.98\textwidth]{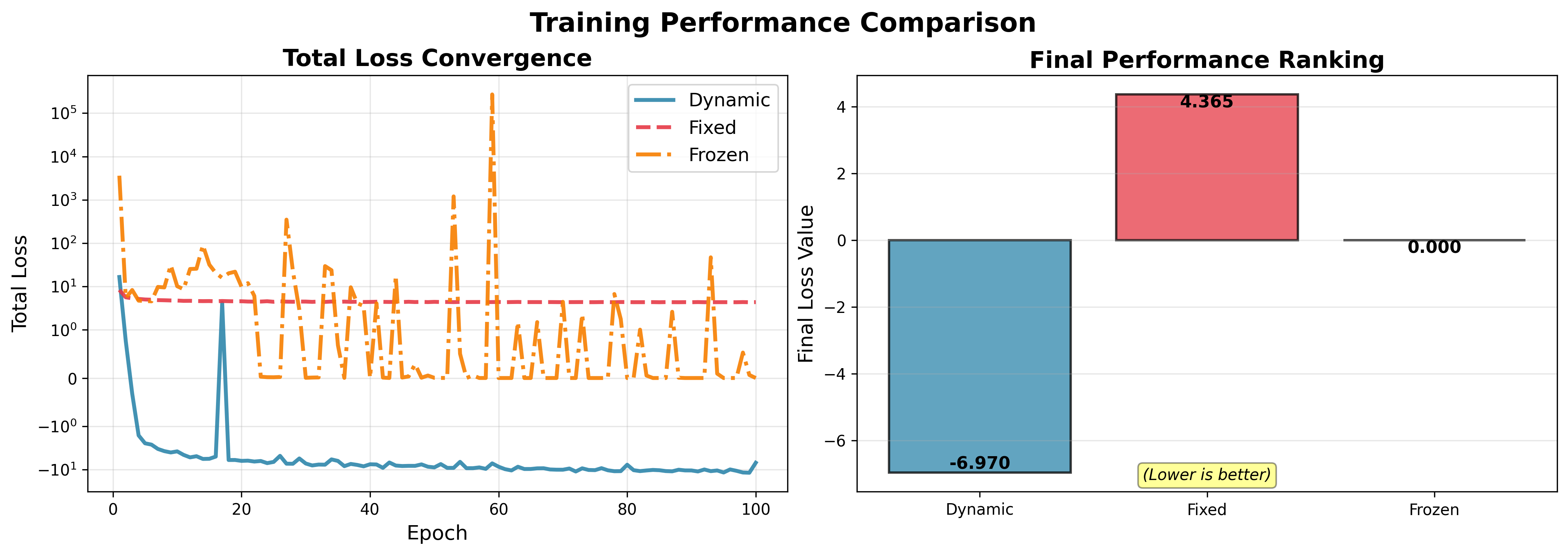}
    \caption[Training performance comparison between weighting regimes]{Proposed dynamic weighting versus fixed and frozen baselines. Left: total loss convergence (log-scale) over $100$ epochs. Right: final loss.}
    \label{fig:dynamic_training_performance}
    \vspace{-0.5\baselineskip}
\end{figure*}

\begin{figure*}[t]
    \vspace{-0.5\baselineskip}
    \centering
    \includegraphics[width=0.98\textwidth]{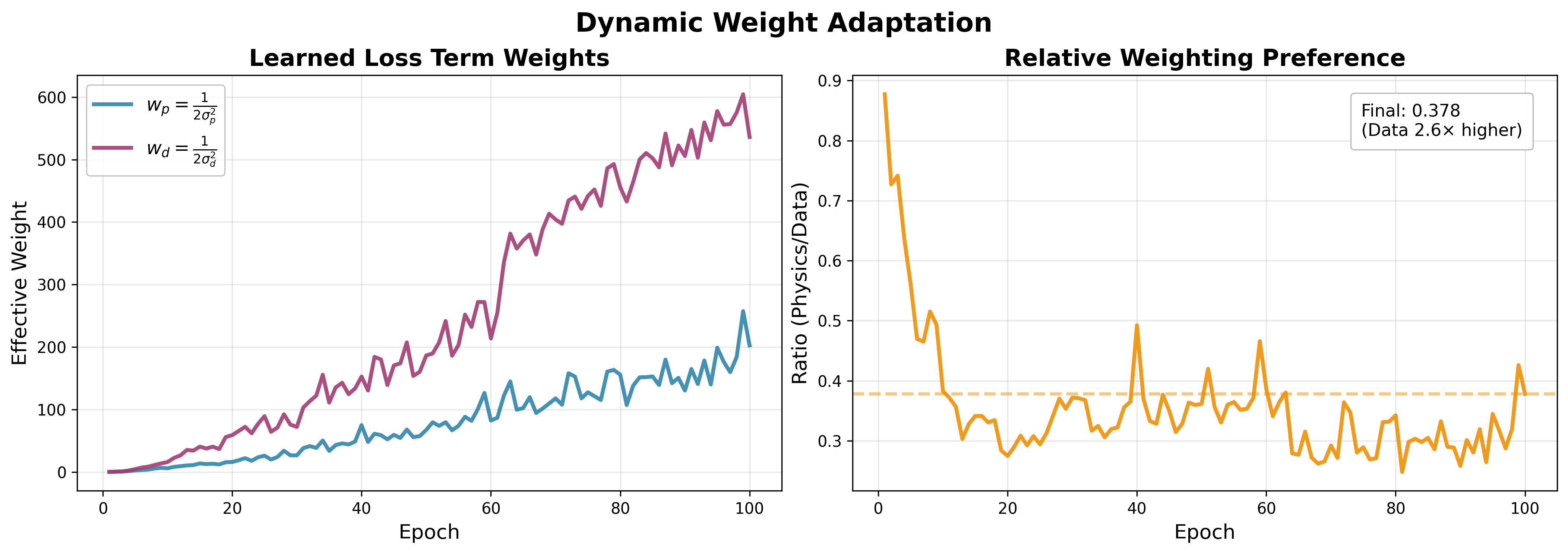}
    \caption[Dynamic loss weights evolution]{Dynamic weight evolution: trajectories of $w_{\text{p}}$ and $w_{\text{d}}$, and ratio $w_{\text{p}}/w_{\text{d}}$.}
    \label{fig:dynamic_weight_adaptation}
    \vspace{-1\baselineskip}
\end{figure*}

\section{Experimental Setup and Results}\label{s:exp_results}
\subsection{Experimental Setup and Model Training}
The FDIA pipeline generates datasets on the IEEE~118-bus system using the optimization process in Section~\ref{ss:attack_construct}. Case data come from \textit{PyPower}, while \textit{Pyomo}~\cite{bynum2021pyomo} formulates attack instances and \textit{IPOPT}~\cite{biegler2009large} solves them. Training uses $14{,}822$ steady-state snapshots with Gaussian measurement noise. Evaluation uses six attack-zone test datasets (two constraint regimes across three zones), with $\sim3{,}700$ samples per dataset. The analysis reports three representative attack areas: Zone~$1$ (buses 10--19, 29, 30, 32, 33, 36, and 112), Zone~$2$ (buses 41, 44, 46--50, 53, 55, 56, 65, and 68), and Zone~$3$ (buses 91, 93, 97--103, and 105). In all cases, the chosen zone satisfies $\mathcal{Z}=\mathcal{B}_{\mathrm{int}}\cup\mathcal{B}_{\mathrm{bnd}}$. Breadth-first search with maximum hop distance $h_{\max}=2$ generates zone candidates of $n_{\min}=3$ to $n_{\max}=20$ buses.

Shared stealth and feasibility settings are fixed across scenarios. The experiments apply the per-bus residual-threshold screen $\tau_P=0.95\bar{\tau}_P$ and $\tau_Q=0.95\bar{\tau}_Q$, where clean baseline scales $\bar{\tau}_P,\bar{\tau}_Q$ use a minimum of $0.01$. Boundary-bus active-transfer mismatch is limited to $\max(0.03|F_{\mathrm{bnd},i}^0|,0.01)$, i.e., $\varepsilon_{\mathrm{bnd,rel}}=~0.03$ and $\varepsilon_{\mathrm{bnd,abs}}=~0.01$. Voltage magnitudes stay within $[0.95,1.05]$ p.u. and angles within $[-\pi,\pi]$. Zero-injection buses, identified by $|P_i^0|,|Q_i^0|<10^{-6}$, are fixed to $P_i^a=Q_i^a=0$. Zone-level power conservation uses the shared tolerance $\varepsilon_{\mathrm{cons}}=\max(10^{-3}|\sum_{i\in\mathcal{Z}} P_i^0|, 10^{-3})$ for both active and reactive balances. For each snapshot and attack, the estimator receives attacked power injections $[\mathbf{P}^a,\mathbf{Q}^a]$, while error is computed against the uncompromised states $[\mathbf{V}^0,\boldsymbol{\theta}^0]$.

Regime-specific defaults follow the implemented formulations. In the baseline state-distortion regime, active and reactive injections are bounded within $\pm 75\%$ of a baseline scale defined as $\max(|\text{baseline injection}|,0.01)$, corresponding to $\kappa_P=\kappa_Q=0.75$ and $\delta_P=\delta_Q=0.01$. In the residual-profile-matching regime, $\beta=0.05$ and $\varepsilon_r=10^{-3}$, with per-bus tolerances $\max(\beta|r_{P,i}^0|,\varepsilon_r)$ and $\max(\beta|r_{Q,i}^0|,\varepsilon_r)$ computed from clean residual references.

All PINN variants share an Optuna-based search procedure and training budget. We compare dynamic uncertainty weighting, fixed static weights, and a frozen variant that reuses final dynamic weights to test whether online adaptation matters during training. Optuna searches $N_{\mathrm{layers}}\in\{2,4,6\}$, neuron width from $64$ to $4096$, batch size from $32$ to $128$, and learning rate in $[10^{-5},10^{-3}]$ (log-uniform), with \textit{swish} activation throughout. Dynamic and fixed variants both tune the ratio-penalty coefficient over $[10^{-4},10]$, and fixed also searches per-component log-sigmas over $[-5,5]$.

Additional constants are $[s_{\min},s_{\max}]=[-4,2]$, $\varepsilon_{\mathrm{norm}}=10^{-8}$, and $\varepsilon_{\mathrm{ratio}}=10^{-12}$. Each model trains for 100 epochs, with performance reported as MAE on voltage magnitudes and angles over all buses. Training uses only clean steady-state operating points, so FDIA tests are out-of-distribution, and minibatch statistics stayed stable.

Each attacked sample maps uniquely to its clean source using state entries outside the manipulated region. The estimator receives attacked injections, while its voltage and angle estimates are evaluated against this uncompromised state. Voltage and angle errors are averaged over all buses and samples, and overall MAE weights both equally. Percentile metrics use the corresponding per-sample errors. Results are averaged equally across zones and then regimes to avoid bias from small sample-count differences.

An ablation study isolates the effect of dynamic weighting by comparing three regimes under identical architectures, optimizer settings, and training data. Our proposed variant is Dynamic uncertainty weighting, which initializes the four per-component log-uncertainties $(s_p,s_q,s_v,s_{\theta})$ to $0.0$ and updates them online during training. The fixed regime holds effective weights constant and tunes them offline with Optuna Tree-structured Parzen Estimator (TPE) to provide a strong static data/physics tradeoff baseline. The frozen regime uses the dynamic run's final uncertainty values and holds them fixed in a separate run, isolating whether adaptation during training matters beyond the final weights.

\subsection{Ablation Study}
Figure~\ref{fig:dynamic_training_performance} shows total-loss trajectories and final values. Dynamic and frozen configurations both converge below the fixed-weight baseline, but the frozen regime is erratic. The dynamic regime converges more smoothly and reaches the lowest final loss, showing that online adaptation is more effective than preset physics/data balances.

Figure~\ref{fig:dynamic_weight_adaptation} shows the weight evolution: the physics term dominates for the first $\sim20$ epochs, after which the data term takes over while physics regularizes, and the ratio $w_{\text{p}}/w_{\text{d}}$ stabilizes around $0.38$ rather than a forced $1{:}1$ balance.

Aggregated over the two FDIA regimes and three zones, dynamic weighting lowers overall MAE against the uncompromised state by $55\%$ versus fixed weighting and $72\%$ versus frozen; per regime, the reduction versus fixed is $38\%$ under baseline state distortion and $61\%$ under residual-profile matching. All variants share the same Optuna trial budget, so the comparison isolates the weighting scheme.

\begin{figure*}[t]
    \centering
    \includegraphics[width=0.48\textwidth]{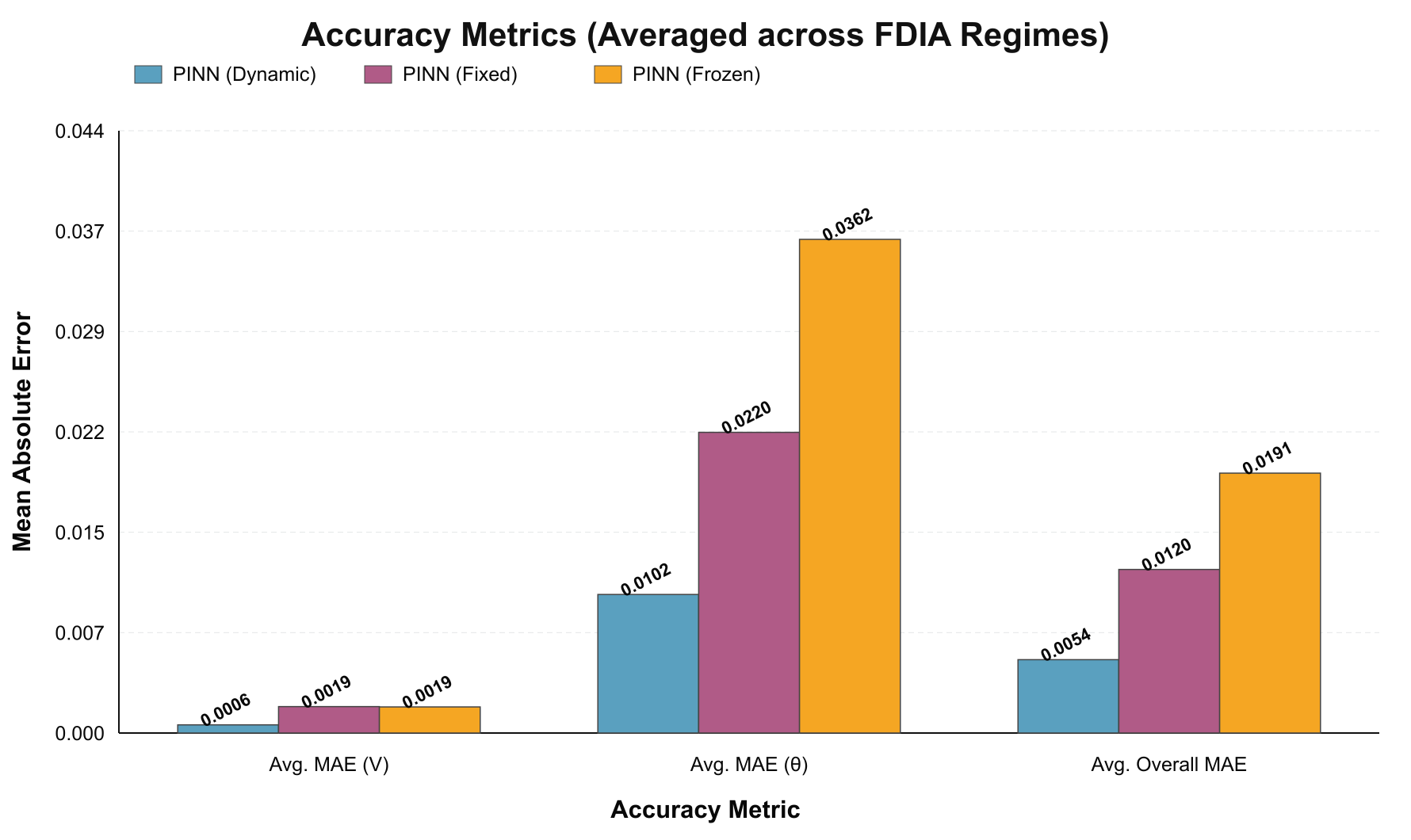}\hfill
    \includegraphics[width=0.48\textwidth]{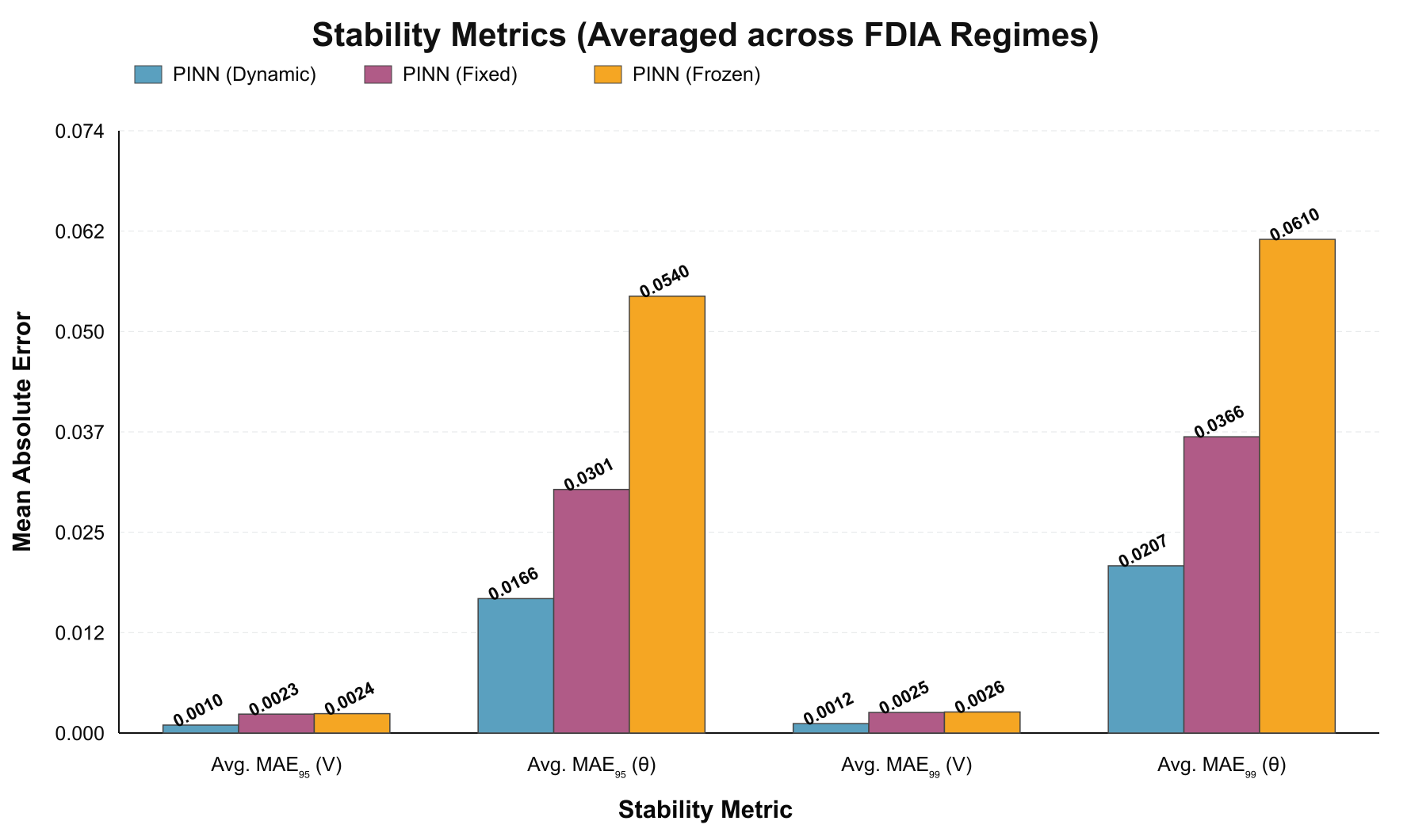}
    \vspace{-0.5\baselineskip}
    \caption[Ablation Study Accuracy and Stability Metrics]{Ablation across weighting regimes. Left: average MAE across $[\mathbf{V},\boldsymbol{\theta}]$. Right: worst-case ($95$th/$99$th-percentile) MAE across $[\mathbf{V},\boldsymbol{\theta}]$.}
    \label{fig:accuracy_metrics_fdia}
    \vspace{-1.5\baselineskip}
\end{figure*}

\begin{table}[t]
    \centering
    \caption[Average MAE across FDIA regimes]{Average MAE across FDIA regimes and zones}
    \label{tab:fdia_mae_by_zone}
    \renewcommand{\arraystretch}{0.92}
    \resizebox{\columnwidth}{!}{
        \begin{tabular}{llccc}
            \hline
            \textbf{Regime} & \textbf{Zone} & \textbf{Dynamic} & \textbf{Fixed} & \textbf{Frozen}\\
            \hline
            Baseline state distortion & 1 & $2.686\times10^{-3}$ & $4.651\times10^{-3}$ & $6.702\times10^{-3}$\\
                                      & 2 & $6.761\times10^{-3}$ & $8.101\times10^{-3}$ & $1.325\times10^{-2}$\\
                                      & 3 & $2.481\times10^{-3}$ & $6.359\times10^{-3}$ & $1.522\times10^{-2}$\\
            Residual-profile matching & 1 & $1.054\times10^{-2}$ & $1.765\times10^{-2}$ & $1.608\times10^{-2}$\\
                                      & 2 & $5.982\times10^{-3}$ & $1.941\times10^{-2}$ & $1.126\times10^{-2}$\\
                                      & 3 & $3.828\times10^{-3}$ & $1.579\times10^{-2}$ & $5.183\times10^{-2}$\\
            \hline
        \end{tabular}
    }
    \vspace{-0.5\baselineskip}
\end{table}

Figure~\ref{fig:accuracy_metrics_fdia} separates these gains by state variable. On average (left), dynamic weighting reduces $V$ MAE by $68\%$ and $\theta$ MAE by $54\%$ relative to fixed weighting; at the tail (right), it reduces the $95$th/$99$th-percentile $V$ MAE by $57\%$/$52\%$ and $\theta$ MAE by $45\%$/$43\%$, so the advantage holds for worst-case as well as average error.

\subsection{Robustness Across FDIA Benchmarks}
Table~\ref{tab:fdia_mae_by_zone} reports overall MAE against the uncompromised state for every evaluated regime--zone pair. It complements the aggregate comparisons in Fig.~\ref{fig:accuracy_metrics_fdia} by exposing the geographic variation hidden by zone averages.
Dynamic weighting has the lowest MAE in all six regime--zone pairs, with zone-wise reductions versus fixed weighting of $17$--$61\%$ under baseline state distortion and $40$--$76\%$ under residual-profile matching, so the advantage is not produced by a single favorable attack area. Tightening the residual constraints also degrades it most gracefully: its zone-average MAE rises by $71\%$, against $177\%$ and $125\%$ for the fixed and frozen variants. The hardest region shifts with the regime---Zone~2 under baseline distortion, Zone~1 under residual matching---underscoring the value of reporting all three zones.

Table~\ref{tab:fdia_benchmark_csr} tests whether the improvement extends beyond the fixed/frozen ablation. The earlier, IEEE~14-trained PINN in~\cite{falas2025csr} maps power injections to voltage/angle estimates via a normalized data loss plus an AC physics loss under a fixed, Optuna-tuned weight. It retrains that architecture on the IEEE~118 dataset and evaluates it on the same benchmarks as the proposed model, which gives lower average MAE under both regimes.
% Table~\ref{tab:scaled_manipulation} and its prose commented out:
% Table~\ref{tab:scaled_manipulation} reverses the test, applying~\cite{falas2025csr}'s single-bus scaled-measurement attack to the proposed model; the published IEEE~14 values are historical context only, since the network differs, so the controlled comparison is between the proposed model's own IEEE~118 one-bus and ten-bus results, which show only a mild error increase as the attack extends to ten buses.

\begin{table}[t]
    \centering
    \caption[Average MAE versus a retrained prior PINN]{Average MAE vs.\ a retrained~\cite{falas2025csr}}
    \label{tab:fdia_benchmark_csr}
    \vspace{-0.5\baselineskip}
    \renewcommand{\arraystretch}{0.9}
    \begin{tabular}{lcc}
        \hline
        \textbf{Model} & \textbf{Baseline} & \textbf{Residual}\\
         & \textbf{Distortion} & \textbf{Matching}\\
        \hline
        Dynamic PINN (proposed) & $3.976\times10^{-3}$ & $6.782\times10^{-3}$\\
        Fixed-weight~\cite{falas2025csr} & $1.40\times10^{-2}$ & $4.51\times10^{-2}$\\
        \hline
    \end{tabular}
    \vspace{-0.5\baselineskip}
\end{table}

% Table~\ref{tab:scaled_manipulation} (Resilience under scaled data-manipulation attacks) commented out:
\iffalse
\begin{table}[t]
    \centering
    \caption[PINN resilience under scaled data-manipulation attacks]{Resilience under scaled data-manipulation attacks}
    \label{tab:scaled_manipulation}
    \renewcommand{\arraystretch}{0.9}
    \begin{tabular}{cccc}
        \hline
        \textbf{Pert.} & \textbf{MAE} & \textbf{MAE} & \textbf{MAE}\\
        \textbf{Level} & \textbf{(\cite{falas2025csr}, 1 bus)} & \textbf{(Ours, 1 bus)} & \textbf{(Ours, 10 buses)}\\
        \hline
        $5\%$  & $7.07\times 10^{-3}$ & $1.140\times 10^{-3}$ & $1.879\times 10^{-3}$\\
        $10\%$ & $1.40\times 10^{-2}$ & $1.387\times 10^{-3}$ & $2.012\times 10^{-3}$\\
        $20\%$ & $2.79\times 10^{-2}$ & $1.302\times 10^{-3}$ & $3.539\times 10^{-3}$\\
        $30\%$ & $4.18\times 10^{-2}$ & $1.776\times 10^{-3}$ & $4.812\times 10^{-3}$\\
        \hline
    \end{tabular}
\end{table}
\fi

%% file: 06_conclusion.tex
\section{Conclusion}\label{s:conclusion}
This work presents a dynamically weighted PINN framework for state estimation under FDIAs. The homoscedastic uncertainty objective learns the data/physics balance during training, avoiding manual loss-weight tuning. On the IEEE~118-bus system, the dynamic variant reduces uncompromised-state overall MAE by $55\%$ relative to fixed weighting and by $72\%$ relative to frozen weighting across baseline state-distortion and stricter residual-profile-matching regimes. The gains persist in both average and high-percentile voltage and angle errors across all three evaluated zones. These results show that adapting the physics/data balance during clean-only training improves robustness as stealth constraints on an unseen AC-consistent state-distortion attack tighten.

\section*{Acknowledgements}
This work is supported by the TwinEU project, which has received funding from the European Union’s Horizon Europe research and innovation programme under Grant Agreement No. 101136119, and by the KIOS Center of Excellence (KIOS CoE), which has received funding from the European Union’s Horizon 2020 research and innovation programme under Grant Agreement No. 739551 and from the Government of the Republic of Cyprus through the Cyprus Deputy Ministry of Research, Innovation and Digital Policy.

%% file: publications.bib
@IEEEtranBSTCTL{BSTcontrol,
  CTLdash_repeated_names = "no"
}

@INPROCEEDINGS{falas2025csr,
  author={Falas, Solon and Asprou, Markos and Konstantinou, Charalambos and Michael, Maria K.},
  booktitle={2025 IEEE International Conference on Cyber Security and Resilience (CSR)}, 
  title={Data Manipulation Attack Mitigation in Power Systems Using Physics-Informed Neural Networks}, 
  year={2025},
  volume={},
  number={},
  pages={693-698},
  doi={10.1109/CSR64739.2025.11129987}}

@ARTICLE{falas2025tii,
  author={Falas, Solon and Asprou, Markos and Konstantinou, Charalambos and Michael, Maria K.},
  journal={IEEE Transactions on Industrial Informatics}, 
  title={Robust Power System State Estimation Using Physics-Informed Neural Networks}, 
  year={2025},
  volume={21},
  number={10},
  pages={8057-8067},
  doi={10.1109/TII.2025.3582293}}


%% file: sources.bib
@article{biegler2009large,
  title   = {Large-scale nonlinear programming using IPOPT: An integrating framework for enterprise-wide dynamic optimization},
  journal = {Computers \& Chemical Engineering},
  volume  = {33},
  number  = {3},
  pages   = {575-582},
  year    = {2009},
  note    = {Selected Papers from the 17th European Symposium on Computer Aided Process Engineering held in Bucharest, Romania, May 2007},
  issn    = {0098-1354},
  doi     = {https://doi.org/10.1016/j.compchemeng.2008.08.006},
  author  = {L.T. Biegler and V.M. Zavala}
}

@book{bynum2021pyomo,
  title     = {Pyomo --- Optimization Modeling in Python},
  author    = {Bynum, Michael L. and Hackebeil, Gabriel A. and Hart, William E. and Laird, Carl D. and Nicholson, Bethany L. and Siirola, John D. and Watson, Jean-Paul and Woodruff, David L.},
  edition   = {3rd},
  series    = {Springer Optimization and Its Applications},
  volume    = {67},
  year      = {2021},
  publisher = {Springer},
  address   = {Cham, Switzerland},
  isbn      = {978-3-030-68927-8},
  doi       = {10.1007/978-3-030-68928-5}
}

@inproceedings{cipolla2018multi,
  author    = {Kendall, Alex and Gal, Yarin and Cipolla, Roberto},
  title     = {Multi-Task Learning Using Uncertainty to Weigh Losses for Scene Geometry and Semantics},
  booktitle = {Proceedings of the IEEE Conference on Computer Vision and Pattern Recognition (CVPR)},
  month     = {June},
  year      = {2018}
}

@article{deng2016false,
  author   = {Deng, Ruilong and Xiao, Gaoxi and Lu, Rongxing and Liang, Hao and Vasilakos, Athanasios V.},
  journal  = {IEEE Transactions on Industrial Informatics},
  title    = {False Data Injection on State Estimation in Power Systems—Attacks, Impacts, and Defense: A Survey},
  year     = {2017},
  volume   = {13},
  number   = {2},
  pages    = {411-423},
  doi      = {10.1109/TII.2016.2614396}
}

@article{ding2020secure,
  author   = {Ding, Derui and Han, Qing-Long and Ge, Xiaohua and Wang, Jun},
  journal  = {IEEE Transactions on Systems, Man, and Cybernetics: Systems},
  title    = {Secure State Estimation and Control of Cyber-Physical Systems: A Survey},
  year     = {2021},
  volume   = {51},
  number   = {1},
  pages    = {176-190},
  doi      = {10.1109/TSMC.2020.3041121}
}

@inproceedings{gao2015cdc,
  author    = {Gao, Sicun and Xie, Le and Solar-Lezama, Armando and Serpanos, Dimitrios and Shrobe, Howard},
  booktitle = {2015 54th IEEE Conference on Decision and Control (CDC)},
  title     = {Automated vulnerability analysis of AC state estimation under constrained false data injection in electric power systems},
  year      = {2015},
  volume    = {},
  number    = {},
  pages     = {2613-2620},
  doi       = {10.1109/CDC.2015.7402610}
}

@inproceedings{iranpour2024fdia,
  author    = {Iranpour, Mohammadreza and Narimani, Mohammad Rasoul},
  booktitle = {2024 56th North American Power Symposium (NAPS)},
  title     = {AC False Data Injection Attacks in Power Systems: Design and Optimization},
  year      = {2024},
  volume    = {},
  number    = {},
  pages     = {1-6},
  doi       = {10.1109/NAPS61145.2024.10741834}
}

@inproceedings{jin2019cdc,
  author    = {Jin, Ming and Molybog, Igor and Mohammadi-Ghazi, Reza and Lavaei, Javad},
  booktitle = {2019 IEEE 58th Conference on Decision and Control (CDC)},
  title     = {Towards Robust and Scalable Power System State Estimation},
  year      = {2019},
  volume    = {},
  number    = {},
  pages     = {3245-3252},
  doi       = {10.1109/CDC40024.2019.9030243}
}

@article{ngo2024physics,
  title    = {Physics-informed graphical neural network for power system state estimation},
  journal  = {Applied Energy},
  volume   = {358},
  pages    = {122602},
  year     = {2024},
  issn     = {0306-2619},
  doi      = {https://doi.org/10.1016/j.apenergy.2023.122602},
  author   = {Quang-Ha Ngo and Bang L.H. Nguyen and Tuyen V. Vu and Jianhua Zhang and Tuan Ngo}
}

@article{raissi2019physics,
  title    = {Physics-informed neural networks: A deep learning framework for solving forward and inverse problems involving nonlinear partial differential equations},
  journal  = {Journal of Computational Physics},
  volume   = {378},
  pages    = {686-707},
  year     = {2019},
  issn     = {0021-9991},
  doi      = {https://doi.org/10.1016/j.jcp.2018.10.045},
  author   = {M. Raissi and P. Perdikaris and G.E. Karniadakis}
}

@article{ran2026correlation,
  author  = {Ran, Xiaohong and Bu, Siqi and Liu, Yangsheng},
  journal = {IEEE Transactions on Smart Grid},
  title   = {Correlation-Exploiting False Data Injection Attack and Corresponding DRL-Based Detection Countermeasure in Partially Observable Grids},
  year    = {2026},
  doi     = {10.1109/TSG.2026.3694065}
}

@article{tian2020hybrid,
  author   = {Tian, Guanyu and Zhou, Qun and Birari, Rahul and Qi, Junjian and Qu, Zhihua},
  journal  = {IEEE Transactions on Neural Networks and Learning Systems},
  title    = {A Hybrid-Learning Algorithm for Online Dynamic State Estimation in Multimachine Power Systems},
  year     = {2020},
  volume   = {31},
  number   = {12},
  pages    = {5497-5508},
  doi      = {10.1109/TNNLS.2020.2968486}
}

@inproceedings{tran2021enhancement,
  author    = {Tran, Minh-Quan and Zamzam, Ahmed S. and Nguyen, Phuong H.},
  booktitle = {2021 IEEE Madrid PowerTech},
  title     = {Enhancement of Distribution System State Estimation Using Pruned Physics-Aware Neural Networks},
  year      = {2021},
  volume    = {},
  number    = {},
  pages     = {1-5},
  doi       = {10.1109/PowerTech46648.2021.9494950}
}

@article{wang2020physicsguided,
  author   = {Wang, Lei and Zhou, Qun and Jin, Shuangshuang},
  journal  = {Journal of Modern Power Systems and Clean Energy},
  title    = {Physics-guided Deep Learning for Power System State Estimation},
  year     = {2020},
  volume   = {8},
  number   = {4},
  pages    = {607-615},
  doi      = {10.35833/MPCE.2019.000565}
}

@article{wang2021understanding,
  author   = {Wang, Sifan and Teng, Yujun and Perdikaris, Paris},
  title    = {Understanding and Mitigating Gradient Flow Pathologies in Physics-Informed Neural Networks},
  journal  = {SIAM Journal on Scientific Computing},
  volume   = {43},
  number   = {5},
  pages    = {A3055-A3081},
  year     = {2021},
  doi      = {10.1137/20M1318043},
  eprint   = {https://doi.org/10.1137/20M1318043}
}

@article{yang2022data,
  author   = {Yang, Qiuling and Sadeghi, Alireza and Wang, Gang},
  journal  = {IEEE Journal on Emerging and Selected Topics in Circuits and Systems},
  title    = {Data-Driven Priors for Robust PSSE via Gauss-Newton Unrolled Neural Networks},
  year     = {2022},
  volume   = {12},
  number   = {1},
  pages    = {172-181},
  doi      = {10.1109/JETCAS.2022.3142051}
}

@article{zhang2019real,
  author   = {Zhang, Liang and Wang, Gang and Giannakis, Georgios B.},
  journal  = {IEEE Transactions on Signal Processing},
  title    = {Real-Time Power System State Estimation and Forecasting via Deep Unrolled Neural Networks},
  year     = {2019},
  volume   = {67},
  number   = {15},
  pages    = {4069-4077},
  doi      = {10.1109/TSP.2019.2926023}
}
